\documentclass[superscriptaddress,longbibliography,twocolumn]{revtex4}
\usepackage[utf8]{inputenc}
\usepackage{mleftright}
\usepackage{graphicx}%
\usepackage{bm}%
\usepackage{graphicx}%
\usepackage{xcolor}
\usepackage[version=4]{mhchem}

\usepackage{fancyvrb}
\usepackage{listings}

\usepackage{pdfpages}

\definecolor{backcolour}{rgb}{0.95,0.95,0.92}
\definecolor{commentgrey}{rgb}{0.4,0.4,0.4}
\definecolor{deepblue}{rgb}{0,0.6,1.0}
\definecolor{lightgrey}{rgb}{0.99,0.99,0.99}
\definecolor{deepred}{rgb}{0.0,0.6,1.0}
\definecolor{deepgreen}{rgb}{1.0,0.0,0.6}
\lstdefinestyle{customstyle}{
    commentstyle=\color{commentgrey},
    basicstyle=\ttfamily\footnotesize,
    breakatwhitespace=false,         
    breaklines=true,                 
    keepspaces=true,                 
    showspaces=false,                
    showstringspaces=false,
    backgroundcolor=\color{lightgrey},
    frame=single,
    showtabs=false,                  
    tabsize=2,
    keywordstyle=\color{deepblue},
    emph={_fit,_map,RandomDescriptor,Module,StructureInput,TopologicalFP,KernelDot,KernelRidge,BayesianHyper,Hyper},
    emphstyle=\color{deepred},   
    stringstyle=\color{deepgreen},
}
\usepackage{hyperref}
\usepackage{acronym}
\usepackage{comment}
\usepackage{siunitx}
\usepackage{url}
\usepackage{graphicx}

\newcommand{\ba}{\mathbf{\alpha}}
\newcommand{\bw}{\mathbf{w}}
\newcommand{\bx}{\mathbf{x}}
\newcommand{\bX}{\mathbf{X}}
\newcommand{\bk}{\mathbf{k}}
\newcommand{\bK}{\mathbf{K}}
\newcommand{\bI}{\mathbf{I}}
\newcommand{\by}{\mathbf{y}}

\begin{document}

\title{BenchML:
an extensible pipelining framework for benchmarking\\ representations of materials and molecules at scale}

\author{Carl Poelking}
\email{carl.poelking@astx.com} 
\affiliation{Astex Pharmaceuticals, Cambridge, UK}
\affiliation{Department of Chemistry, University of Cambridge, UK}

\author{Felix A. Faber}
\affiliation{Department of Physics, University of Cambridge, UK}

\author{Bingqing Cheng}
\email{bingqing.cheng@ist.ac.at} 
\affiliation{The Institute of Science and Technology Austria, Am Campus 1, 3400 Klosterneuburg, Austria}

\begin{abstract}
We introduce a machine-learning (ML) framework for high-throughput
benchmarking of diverse representations of chemical systems against datasets of materials and molecules. The guiding principle underlying the benchmarking approach is to evaluate raw descriptor performance by limiting model complexity to simple regression schemes while enforcing best ML practices, allowing for unbiased hyperparameter optimization, and assessing learning progress through learning curves along series of synchronized train-test splits. The resulting models are intended as baselines that can inform future method development, next to indicating how easily a given dataset can be learnt. Through a comparative analysis of the training outcome across a diverse set of physicochemical, topological and geometric representations, we glean insight into the relative merits of these representations as well as their interrelatedness.
\end{abstract}

\maketitle

\section{Introduction}

Making accurate predictions of materials and molecular properties while using minimal computing and experimental resources continues to be a grand challenge in the chemical sciences. Machine learning (ML) has emerged as a promising tool to address this challenge by performing statistical learning on relatively few data points, and then inferring the properties of new examples. In the past decade, ML methods for chemistry have yielded remarkable accuracy for a wide array of materials properties -- from atomization energies, to forces, spectra, stability, optical properties, drug activities and many more~\cite{haghighatlari2020learning,tkatchenko2020machine,vonLilienfeld2020,behler2021four, keith2021combining, Deringer2019}.

Broadly speaking, an ML regression model of a chemical system operates in two key stages: First, translating the data samples (i.e., molecules or materials) into appropriate mathematical representations; second, applying a regression algorithm to these representations. ML for chemistry is thus somewhat different from many traditional ML tasks in natural language or image processing where the input tensors are usually well-defined -- i.e., there is a meaningful ``native'' representation of the input data that is by and large unambiguous. Driven by the observation that the choice of representation plays an outcome-determining role in a chemical model~\cite{vonLilienfeld2020}, a large range of (competing) representations have been developed for describing materials and molecules, complemented by neural-network based approaches that learn their representations on the fly~\cite{musil2021physics}.

\begin{figure*}[hbt]
\includegraphics[width=1.0\textwidth]{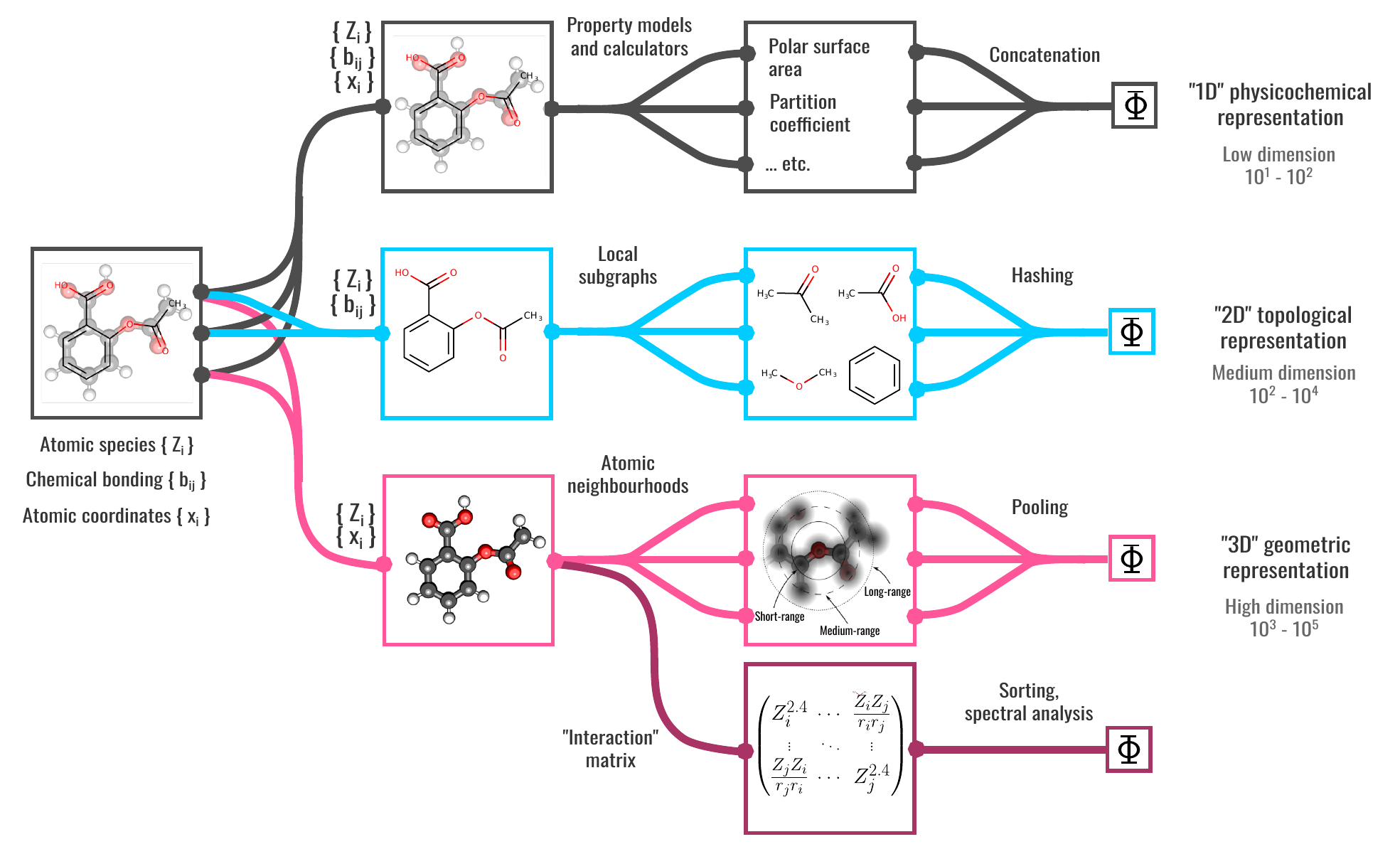}
\caption{ Taxonomy of atomic and molecular representations. We distinguish between: 1D physicochemical representations, 2D topological representations, and 3D geometric representations. The latter includes two subtypes: Atomic-neighbourhood-based and interaction-matrix-based descriptions. The descriptor families differ in particular with respect to: (i) the structural information they use (atomic identities, chemical bonding, atomic coordinates); (ii) how they process the structure (property calculators, graph partitioning, neighbourhood expansions); and (iii) how they form a global representation (hashing, pooling, etc.). Their dimensions can also differ drastically, with 1D physicochemical fingerprints typically being relatively compact ($d = 10^1 - 10^2$), and geometric representations usually high-dimensional, in particular for multi-element structures ($d = 10^3 - 10^5$).
}

\label{fig:representations}
\end{figure*}

At a coarse level, the representations of chemical systems we consider fall into three main categories (Fig.~\ref{fig:representations}): (i) physical- or chemical-property-based ``1D'' representations, incorporating, e.g., measures of polar surface area or lipophilicity~\cite{ertl2000fast,wildman1999prediction}, (ii) topological ``2D'' fingerprints such as the ECFP family of fingerprints~\cite{rogers2010extended}, and (iii) atomic-coordinate-based ``3D'' representations, such as the Coulomb Matrix (CM)\cite{rupp2012fast}, Smooth Overlap of Atomic Positions (SOAP)~\cite{BartokGabor_Descriptors2013} or Atom-Centered Symmetry Functions (ACSF)\cite{behler2011atom}). For each choice of representation, there is typically still significant freedom (i.e., complexity) in assembling its components and selecting its hyperparameters. For physicochemical representations this freedom consists in particular of what system properties to include in the final set of features. For 2D topological fingerprints, parameters such as the topological radius as well as, at a lower level, the hash function itself may need to be customized. For 3D representations, basis functions and/or length-scale hyperparameters need to be specified, including, in particular, the cutoff that defines the size of an atomic neighbourhood.

  \begin{figure*}[hbt]
\includegraphics[width=1.0\textwidth]{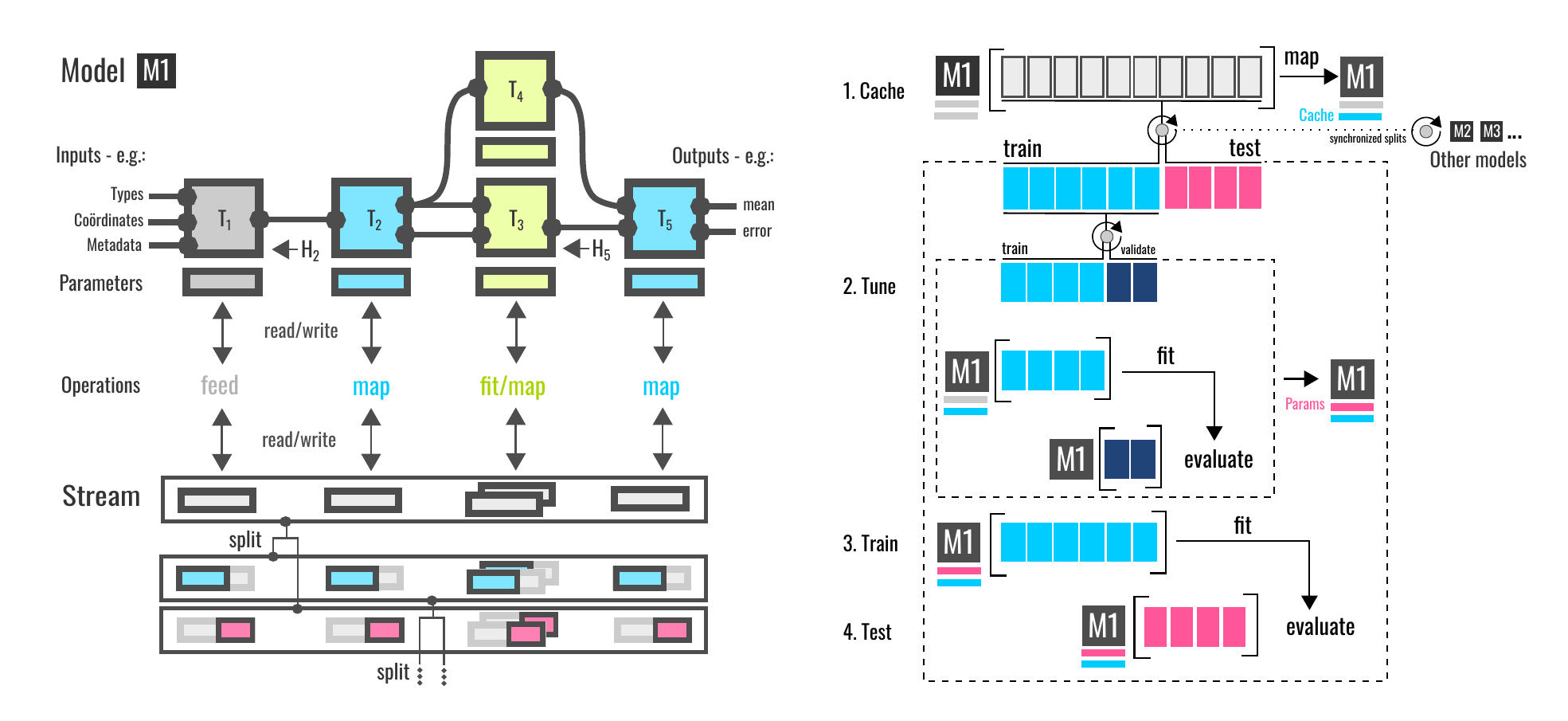}
\caption{
BenchML design schematics. Left: A model BenchML pipeline, consisting of interconnected data transforms T$_i$. Each transform can have several input and ouput channels. Dependency hashes H$_i$ help identify which data objects can be precomputed and cached. The pipeline acts on a data stream from/to which the transforms read/write in a controlled fashion. The streams are used for storing precomputed data (such as kernel or design matrices), which are sliced appropriately during splits. Right: Benchmarking workflow with four stages: Caching (precomputation), tuning (hyperparameter optimization), fitting (with the optimal set of parameters), and testing on withheld test data. Note the coloured rectangles which indicate individual data samples. Angular brackets denote model execution (fitting or mapping). The bars below the model symbol (``M1'') indicate the parametrization and cache state.
}
\label{fig:demo}
\end{figure*}

Building and validating a predictive ML model for chemical systems will typically imply running an objective benchmark in which disparate models compete against each other. This can be much harder than what one might expect: ``Objectivity'' is a tough objective, simply due to subjective choices with regards to dataset compilation, test set generation, and ``favourite'' metrics. The key difficulty, however, lies in investing an equal amount of effort into the fine-tuning of the competing models. In this regard, when faced with a multitude of hyperparameters, some deriving from the representation itself, and others inherited from the predictor, the modeller typically needs to draw a dividing line between hyperparameters that are treated as ``constant'' (having gone through, for example, a manual refinement loop) versus those that are considered ``fluid'' and can thus be dealt with in a nested on-the-fly hyperparameter search. 

Regarding the regression stage, further design choices range from how to normalize the representation (i.e., design) matrix; whether and, if so, how to perform feature selection; how to summarize (``pool'') atomic representations of a system into a single global representation vector; which kernel and kernel normalization to use; as well as whether or how to incorporate regularization and variance reduction techniques. Not all of these choices can be easily treated as model hyperparameters without combinatorially inflating the hyperparameter search and rendering model training cumbersome and expensive.

\begin{figure*}[hbt]
\includegraphics[width=0.97\textwidth]{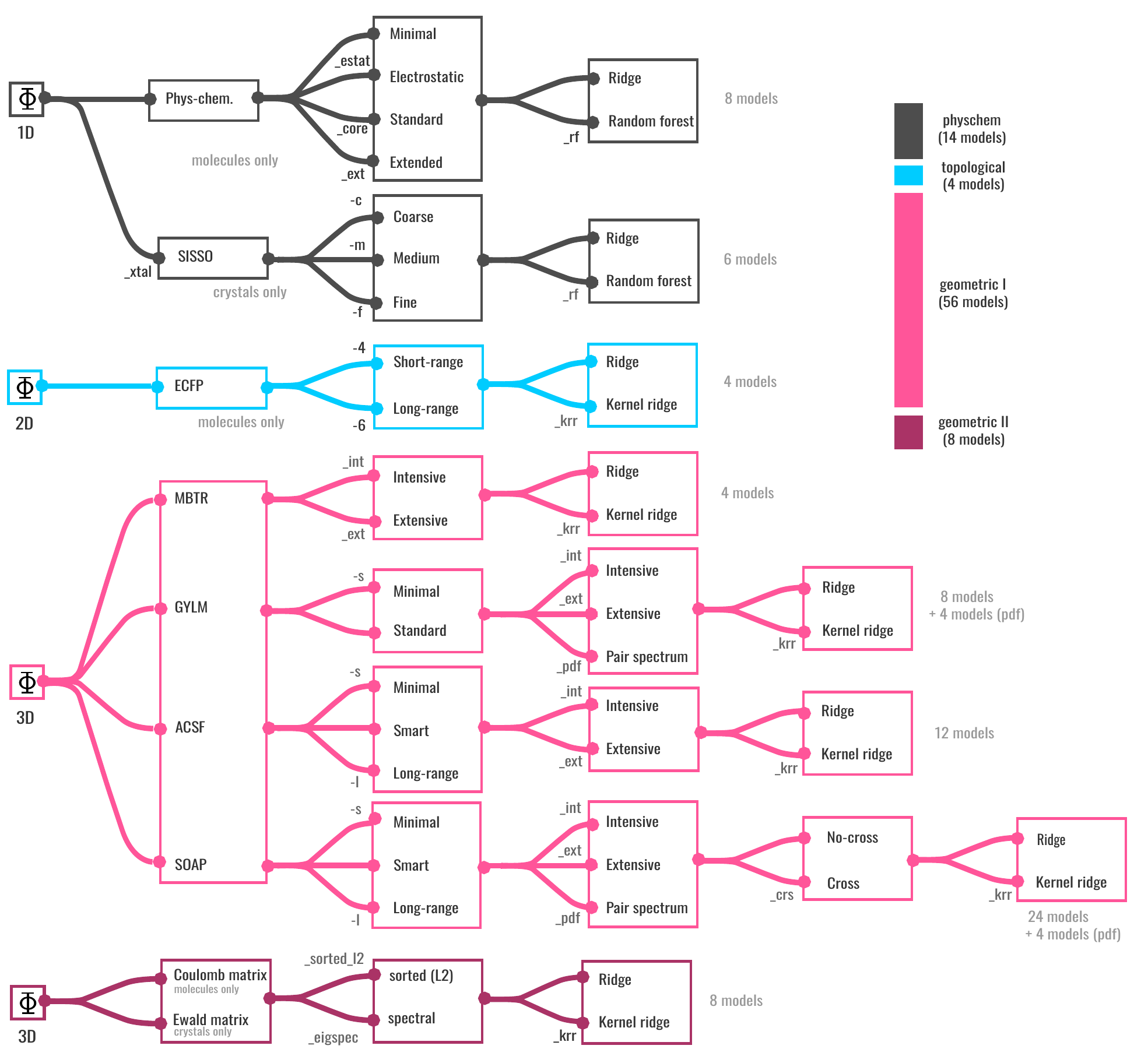}
\caption{ Model phylogeny: ``1D'' physicochemical models (black tree), ``2D'' topological models (blue tree), ``3D'' geometric models (red trees). Next to a descriptor (top level) and regressor (bottom level), each model may additionally specify a descriptor sub-type (minimal, extended, etc.) and a pooling (extensive, intensive, pair correlation) or other reduction rule (sorting, spectral projection). In the pooling category, the ``pair spectrum'' refers to the pair contraction rule in eq.~\ref{eq:pair_spectrum}. For the Coulomb and Ewald matrix, ``sorted (L2)'' and ``spectral'' specify how the interaction matrix is flattened into a single representation. For SOAP, ``no-cross'' vs ``cross'' determines whether element cross-channels in the power spectrum are included. 1D physicochemical representations consider four different subsets of standard molecular properties. The 1D crystal representation on the other hand builds on a subset of the atomic features considered by SISSO~\cite{luca2018sisso}. The branches are annotated with acronyms that are concatenated to produce a unique model tag for later reference (example: {\it soap-l\_int\_krr}, which corresponds to a kernel-ridge model with a long-range SOAP descriptor and intensive global representation).
}

\label{fig:model_phylogeny}
\end{figure*}

In order to compare disparate models on the same footing we need to make the distinction between the merits of the representations themselves, versus the surrounding data pipeline and infrastructure used to embed them in a final ML model. For example, when comparing a model that uses a particular topological representation with a second model that embeds a SOAP descriptor into a kernel ridge regressor using a specific pooling step, it will not be immediately clear whether the observed difference in performance is a result of the choice of representation, the postprocessing and regression algorithms placed thereon, or a combination of the two.

Beyond these confounding factors that complicate the analysis of a benchmark, there are other hazards that can bias the benchmark's outcome. A grey area, for example, is applying standard scalers to the representation matrix before entering a train-test loop, thus leaking information from the test data into the training. Another pitfall concerns non-identical test-train splits when comparing different models, which can randomly but noticeably bias the benchmark results in particular for smaller datasets. 
One more serious and probably not uncommon hazard concerns manual hyperparameter fitting, as partially touched upon above: In order to avoid a combinatorially expensive grid optimization, a modeller may choose to handpick some parameters or manually tune aspects of the overall architecture (including the representation itself) in a trial-and-error fashion, while tracking its performance on a particular dataset. This can result in a model performing particularly well on this specific dataset, due to some model parameters having been ``accidentally'' optimized -- through manual refinement -- across the entire set rather than just a subset.

Finally, additional performance bias may come from the datasets themselves, if, somehow, they allow ML models to take shortcuts and exploit unintended systematic trends -- resulting in ``Clever-Hans'' models ~\cite{pfungst1911clever} that perform well while learning little. As an example, in qm9 (a popular benchmark set in chemistry that contains 13k organic molecules composed of up to nine heavy atoms C, N, O, and F), there is a spurious trend that the atomization energy per atom scales inversely with the total number of atoms~\cite{de+16pccp,cheng2020mapping}. This turns out to be the result of most molecules containing nine heavy atoms, with molecules sampled in a way that those with fewer atoms tend to have more double and triple bonds. This trend may be picked up by an ML model, in which case it would harm the model's ability to generalize to real-world examples once deployed.

Partially because of such intricacies, some doubt has been cast on the viability of ML models in chemical research~\cite{Artrith2021}. As one example, in a study of a Buchwald-Hartwig cross-coupling reaction, the authors reached the conclusion that the combination of physicochemical descriptors and random-forest regression significantly improved predictions of reaction yields~\cite{Ahneman2018}. 
Later, however, it was suggested that the good metrics obtained by the authors were in fact a Clever-Hans-type artifact -- with the model basing its predictions on the presence of certain tell-tale reactants -- and that a similar accuracy could thus be achieved using one-hot molecular encodings or random features~\cite{Chuang2018}. 

To address some of these complexities around building and benchmarking of ML models for materials and molecules, we here introduce BenchML, a machine-learning framework designed to turn benchmarking of chemical representations into a routine task. In essence, BenchML implements a pipelining model that allows us to design and evaluate ML architectures of tunable complexity, and study performance trends across diverse collections of representations and datasets. The framework is highly extensible due to its use of modular data transforms that make up the expression graph of the pipeline. Precomputation of expensive transformations is thus managed automatically in order to speed up the training-test loop as well as any nested grid-based or Bayesian hyperparameter search. We stress that the scope of BenchML extends beyond performing benchmark tests of chemical representations. Next to providing a convenient way to engage with a new, unfamiliar dataset, the ML models of the default BenchML model library are also intended to serve as baselines for more sophisticated, and particularly newly developed, supervised ML models -- including, but not limited to, convolutional neural networks.

This paper is organized as the follows: In Sec.~\ref{sec:method} we introduce the design principle and the architecture of BenchML. In Sec.~\ref{sec:analysis} we demonstrate a specific application with detailed analysis. In Sec.~\ref{sec:examples} we illustrate an example benchmark on some widely used as well as specialized chemical datasets -- covering the prediction of energetics, thermodynamics and reactivity in molecular and crystalline systems.

\section{Methods~\label{sec:method}}

\subsection{Overview of the BenchML framework}

BenchML is a framework designed to address some of the hidden complexities around data-driven materials and molecular modelling, and turn embedding of new representations into general predictors into a routine task. 
From a conceptual point of view, BenchML enables the transition from the low-level ``fit -- predict'' approach (as represented, e.g., by scikit-learn and related libraries) to a higher-level ``build -- benchmark -- deploy'' framework, that allows for releasing robust, finely tuned, well-tested models. In an industrial setting, where ML life-cycle management -- typically referred to as ``MLOps'' -- is crucial, this BenchML workflow can be easily incorporated into any MLOps infrastructure (such as MLFlow~\cite{mlflow2021}) for organization-wide deployment.

BenchML sits on top of lower-level plugin libraries such as {\it dscribe}, {\it asaplib}, or {\it gylmxx} that specialize in a particular set of representations, transformations, regression or filtering techniques. As a key design objective, adding a new data transform that wraps external methods comes with minimal overhead and is achievable with just a few lines of code.

\begin{figure}[t]
\includegraphics[width=0.5\textwidth]{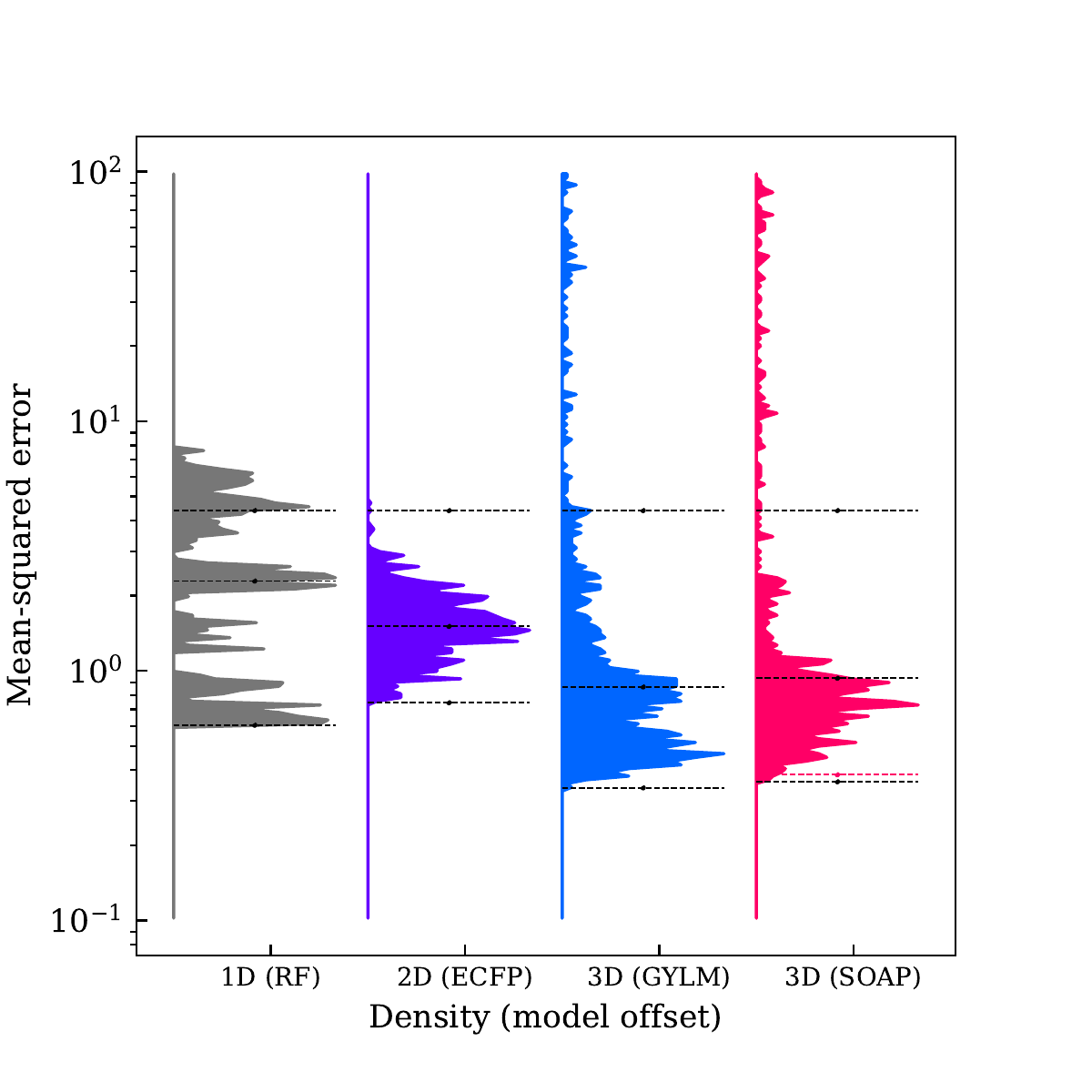}
\caption{ Distribution of the MSE for the aqsol dataset, generated from 1000 randomly drawn hyperparameter settings. The four models, from left to right, are: a physicochemical (1D) random forest, a 2D-topological, and two 3D-convolutional kernels (GYLM and SOAP). Note that the hyperparameter settings are sampled from a large combinatorial space, but that -- importantly -- the regularization strengths of the regressors are tuned for each setting independently using nested splits. For each distribution, the dashed horizontal lines indicate, from top to bottom: the variance of the targets across the entire dataset, the observed model-specific median and minimum of the MSE. %
}
\label{fig:hyper_dist}
\end{figure}

\subsection{The BenchML pipeline}

BenchML follows a simple pipelining concept: Pipelines are directed graphs of data transforms that act on an input data stream, keeping track of dependencies, caching results (where appropriate) and enabling hyperparameter optimization via grid-based and Bayesian techniques. 
The transforms encapsulate a variety of ML methods, from representations, matrix reductions, data filtering, feature selection to regressors, classifiers, ``ensemblizers'' and ``conformalizers'' (the latter take a predictor and turn it into a confidence-calibrated estimator). 
For a more detailed discussion of the transforms implemented to date, we refer the reader to the library's online documentation. 

A schematic representation of the pipelining approach and benchmarking framework is shown in Fig.~\ref{fig:demo}. During the training (``fitting'') and prediction (``mapping'') stage, the data transforms (labelled $T_1$ to $T_5$) read from and write to a data stream that stores intermediate results (such as a normalized design matrix). Importantly, the streams implement data splitting, which is crucial for constructing, tuning and testing ML models efficiently and in a way that prevents cross-contamination. These splits can occur at various stages of model execution and testing. They include: training-test splits to measure prospective performance, training-validation splits to select hyperparameters, training-calibration splits to gauge confidence predictors, and bootstrapped ``splits'' for ensembling and variance estimates.

With splits being so essential, caching and precomputation of data is necessary to be able to train and evaluate models quickly and with minimal computational expense. Consider, for example, a standard benchmarking loop consisting of 100 different train-test splits. For each split, 10-100 different hyperparameter settings are subjected to 10-fold nested validation (Fig.~\ref{fig:demo}). This means that just this single model gets retrained on the order of $10^4$ to $10^5$ times, resulting in a potentially considerable computational cost. Dependency hashing within the BenchML pipelines helps to easily identify which parameters and transforms can be precomputed, and which ones need to be re-evaluated during the hyperparameter search. 

To enable caching and precomputation, when implementing a new transform, all that is required is to specify what data the transform reads from and emits into the stream, as well as the type of that data (see listing~\ref{lst:descriptor} for an example). This informs the pipeline how to process the data when a certain split is applied. Frequently, precomputed fields are either a design matrix or kernel matrix, with the required slicing operations differing between these two types. Finally, beyond annotating the input and output data types of the new transform, what is needed additionally is to specify required and default parameters, and overload the {\it map} and (optionally) {\it fit} operations. The transform can then be incorporated into a pipeline or ``module'', as exemplified in listing~\ref{lst:module}.

\subsection{Representations of chemical systems}

Thus far we have described the pipelining approach behind BenchML, which is of course a general concept and not specific to the modelling of chemical systems. For the purpose of this benchmark, as its core component, each model is based on a single chemical representation of one of the types shown in Fig.~\ref{fig:representations}. An overview of the set of models considered in the benchmark is provided in Fig.~\ref{fig:model_phylogeny}. Here we will discuss in more detail aspects of the model architectures, in particular the different pooling, post-processing and regression rules. For a detailed technical discussion of the representations themselves, we refer the reader to a recent review on the topic~\cite{musil2021physics}. 

\paragraph{Global and atomic representations}

Global as opposed to atomic representations are intended to capture the overall configuration of the whole molecule or bulk material. Some representations are global to begin with -- consider, e.g., Morgan fingerprints that record the presence or absence of specific atomic fragments~\cite{2010_Rogers}), or the Coulomb Matrix, which sequences the pairwise distances between atoms of the structure~\cite{rupp12prl} into a global array. In cases such as these, the global representations can be used as the raw input of BenchML, subject only to potentially different normalization rules, such as the p-norm or feature-wise whitening.

Some representations will, however, describe the system as a set of individual atomic environments, $\mathcal{X}_1, \ldots \mathcal{X}_i \ldots \mathcal{X}_N$, each consisting of the atoms (chemical species and position) contained in their neighbourhood defined by a cutoff radius $r_\mathrm{cut}$ centered around atom $i$. There are many ways how the resulting matrix of atomic descriptors can be reduced into a single molecular vector. In a model pipeline these reduction or pooling rules can be shared across representations and are thus encapsulated in separate data transforms that for atomic representations directly succeed the descriptor calculation step (see the second level of the 3D branch of Fig.~\ref{fig:model_phylogeny}). 

One of the options used by the models in the benchmark is to derive the \emph{intensive} representation for a structure $A$ by averaging over the atomic representations,
\begin{equation}
    \Phi (A) = \dfrac{1}{N_A}
    \sum_{i \in A}^{N_A} \bm{\psi} (\mathcal{X}_i),
\label{eq:fp-global}
\end{equation}
where the sum runs over all $N_A$ atoms $i$ in structure $A$; $\mathcal{X}_i$ is the environment of atom $i$. 
When there are multiple chemical species, the representations for the local environments of different species can either be included in the single sum, or the averaging can be performed for the environments of each species independently, with the final molecular vector obtained by concatenating their averaged local representations. Toggling between these two options can be dealt with as part of the hyperparameter search.

Alternatively, the \emph{extensive} global representation uses
\begin{equation}
        \Phi_\mathrm{ext} (A) = 
    \sum_{i \in A}^{N_A} \bm{\psi} (\mathcal{X}_i).
\end{equation}
For this benchmark, we assume the intensive representation by default. Models using an extensive representation will thus be explicitly annotated with a subscript ${ext}$.

Note that there are several other ways how to construct these global representations -- for example, by using an RMSD-based best match assignment between the environments of separate structures (resulting in an implicit global feature space), or combining local representations using a regularized entropy match (REMatch)~\cite{de+16pccp}. However, next to being computationally expensive, these methods are highly nonlinear adaptations of the underlying descriptor, and are thus beyond the scope of this benchmark study.

As a test, we incorporate one novel way of obtaining the global representation for those atomic descriptors that are based on expansions of the atomic density in terms of spherical harmonics. These descriptors have components $\psi_{nlm}$, where $n$ is now a summary index over radial components and atomic species, and $lm$ indicates the angular momentum channel. A generalization of the SOAP power spectrum then uses non-local contractions over the magnetic quantum number $m$ to arrive at a global representation
\begin{equation}
    \Phi_{nkl}(A) \propto \sum_{i \in A}^{N_A} \sum_{j \in A}^{N_A} \psi_{nlm}(\mathcal{X}_i) \ \psi^*_{klm}(\mathcal{X}_j).
    \label{eq:pair_spectrum}
\end{equation}
This non-local contraction can be interpreted as a generalized form of a pair-distribution function (PDF) that simultaneously captures species, radial and angular cross-correlations. For our benchmark, we have included such non-local extensions for the SOAP~\cite{BartokGabor_Descriptors2013} and GYLM~\cite{poelking2020meaningful} descriptor in our model library (see the {\it pair-spectrum} node in Fig.~\ref{fig:model_phylogeny}). As this PDF contraction changes the representations' behaviour on a basic level, we treat the resulting models as a separate family, referred to in the benchmark of section~\ref{sec:examples} as {\it pdf-soap} and {\it pdf-gylm}. 

\paragraph{Length-scale hyperparameters}

Many atomic representations (e.g. ACSF, SOAP, GYLM) use length-scale hyperparameters that need to be appropriately chosen for a given problem and system. To a limited degree, these hyperparameters can of course be addressed within the hyperparameter search. However, a complete combinatorial sweep is usually expensive given the large set of hyperparameters associated with basis-function-based representations. The computational cost grows further as representations at multiple resolutions are joined together in order to build yet more powerful and flexible models. It is then desirable to use heuristics to automatically select these hyperparameters. The set of heuristics used here has previously been described in Ref.~\citenum{cheng2020mapping}, who based the length-scale hyperparameters for a system with arbitrary chemical composition on characteristic bond lengths estimated by computing a minimal bond length $r_\mathrm{min}^{Z}$ and typical bond length $r_\mathrm{typ}^{Z}$ for each species $Z$ from a set of equilibrium structures with varying coordination numbers. These characteristic scales are finally compiled into a look-up table to be queried at training time.

For the SOAP representation, the standard ``smart'' selection thus involves two sets:
The first SOAP has $r_\mathrm{cut}^1 = \max(1.56 \times  \min_{Z} r_\mathrm{min}^{Z},~2 \AA{})$, which focuses on the shortest length scale of the system.
The second SOAP has $r_\mathrm{cut}^2 = \max(1.56 \times  \max_{Z} r_\mathrm{typ}^{Z}, ~1.2\times r_\mathrm{cut}^1)$,
which is usually large enough to capture at least the second neighbour shell.  
The long-range variant combines two SOAP representations: The first has a shorter range 
$r_\mathrm{cut}^s = \max(2.34\, \min_Z r_\mathrm{min}^Z, 3~\mathrm{\AA})$,
the second a longer range $r_\mathrm{cut}^l = \max(2.34\, \max_Z r_\mathrm{typ}^Z, 1.2\, r_\mathrm{cut}^s)$, both with basis dimensions are $n_\mathrm{max}=8$ and $l_\mathrm{max}=4$. 
The minimal variant includes one representation with a range of 
$r_\mathrm{cut}^l = 1.1\, \max_Z r_\mathrm{typ}^Z$, $n_\mathrm{max}=4$, and $l_\mathrm{max}=3$. 
The Gaussian function width $\sigma$ is always set to $\sigma = r_\mathrm{cut}/8.$

\subsection{The regressors in BenchML}

In the absence of confidence calibration or attribution steps, the regressor serves as the final output node of a model (Fig.~\ref{fig:model_phylogeny}). In principle any regressor, such as a neural network, support vector machine, Gaussian process, etc., can be incorporated into a BenchML pipeline. For the sake of benchmarking representations rather than predictors, however, we here resort to only simple and widely used regressors that allow us to emphasize the raw performance of the underlying representation. We briefly recapitulate ridge regression and kernel ridge regression -- two of the three regression types employed in our benchmark. The third type -- random forest regression, an ensemble technique based on decision trees -- is used only in conjunction with 1D physicochemical fingerprints, and we refer the interested reader to the original paper by Breiman for details~\cite{breiman2001random}.

\paragraph{Ridge regression}

Given $N$ data samples $\{ (y_i,\bx_i)\} $, a linear regression model uses the ansatz
\begin{equation}
    \by = \bX \bw + \mathbf{\epsilon},
\end{equation}
where $\by\equiv (y_1,\cdots,y_N)^T$ is the dependent variable, the design matrix $\bX =(\bx_{1},\cdots,\bx_{N})^T$ is the $d$-dimensional independent or input variable, and $\mathbf{\epsilon}$ is a random variable with zero mean. The coefficients $w_i$ are the parameters of the model. Whereas a traditional least-squares fit is obtained by minimizing the data-dependent square error over all training examples, in ridge regression, this loss function furthermore includes the L2-norm of the parameters $\bw$, thus resulting in a regularized linear fit
\begin{equation}
\mathbf{w}_{ridge}(\lambda)= \underset{\bw\in\mathcal{R}^d}{\operatorname{argmin}} \dfrac{1}{2}||\bX\bw-\by||^2 + \dfrac{\lambda}{2} ||\bw||^2,
\label{eq:ridge}
\end{equation}
with an appropriately chosen regularization strength $\lambda$. We can solve for $\bw$ by equating the gradient of Eqn.~\eqref{eq:ridge} with respect to $\bw$ to zero. This leads us to the closed-form expression for the fit coefficients
\begin{equation}
    \bw = (\bX^T \bX +\lambda \bI)^{-1}
   \bX^T \by,
\end{equation}
where $\bI$ is the $d$-rank identity matrix.

\paragraph{Kernel ridge regression}

Kernel ridge regression (KRR) is the analogue of ridge regression over an implicit feature space induced by a positive semi-definite kernel function $k(\bx_i,\bx_j)$. This function measures the pairwise similarity among data samples. For the purpose of our benchmark, we use a simple dot-product kernel $k(\bx_i,\bx_j) = (\bx_i \bx_j^T)^\nu$ suited for descriptors with positive components $x_{i\alpha} \geq 0$. A positive integer exponent $\nu$ controls the nonlinear degree of the regression. The coefficients $\ba$ of the KRR models are determined using
\begin{equation}
    \ba = -\left( \bK + \lambda \mathbf{I}\right)^{-1} \lambda\by,
\end{equation}
where $\bK$ is the \emph{kernel matrix} with components $\bK_{i,j} = k(\bx_i,\bx_j)$.
For a prospective sample $\bx$, the prediction can be expressed as
\begin{equation}
    f(\bx) = \bk\left( \bK + \lambda \mathbf{I}\right)^{-1}  \by,
\end{equation}
where $\bk= (k(\bx_1, \bx), . . . , k(\bx_n, \bx))$ is the vector of inner products between the training data and the probe $\bx$.

Note that in both ridge and kernel ridge regression, the regularization strength $\lambda$ is a hyperparameter that typically has a major impact on a model's performance. Even though heuristics informed by the data distribution and descriptor characteristics can often be used to select this parameter with adequate accuracy, we here incorporate $\lambda$ into each model's automatic hyperparameter search, sweeping a broad range from $\lambda = 10^{-9}$ to $10^{7}$.

Naturally, even with the regularization strength adjusted optimally, a model's performance will still fluctuate significantly subject to how its other hyperparameters are set. The range of this fluctuation gives some insight into how parametrically ``robust'' a particular model is. Indeed, one of the reasons why random forests and topological fingerprints have established themselves as staple techniques in cheminformatics has to do with their tendency to produce reliable models that are unlikely to yield divergent predictions -- a trait not easily reproduced with more complex geometric representations. To illustrate this, Fig.~\ref{fig:hyper_dist} shows the error distribution measured over a large number of random hyperparameter settings for four models when trained on solubility data (ESOL dataset): A physicochemical random forest regressor, a topological kernel, and two geometric kernels (SOAP and GYLM). The geometric kernels feature heavy tails in their error distribution that are indicative of ``divergent'' models that completely failed to train. These long tails are absent for both the physicochemical and topological framework. Nevertheless, the geometric models display a significantly improved peak performance if their hyperparameters are set adequately. Furthermore, for SOAP, the heuristic rules for basis-function selection outlined above are able to almost precisely pinpoint the optimal setting, as indicated by the dashed red horizontal line.

\begin{table}
\centering
\begin{tabular}{l|ccc}
Activity & RMD & ECFP-SVM & ECFP-SVM            \\
 prediction & (Ref.~\cite{lee2019ligand}) & (Ref.~\cite{lee2019ligand}) & (BenchML)    \\
\hline
MOR1   & 0.99             & 0.70                  & $0.995\pm0.002$                \\
5-HT2B & 0.93             & 0.67                  & $0.979\pm0.004$                \\
ADRA2A & 0.90             & 0.61                  & $0.928\pm0.008$                \\
HistH1 & 0.97             & 0.65                  & $0.987\pm0.003$                \\
hERG   & 0.83             & 0.60                  & $0.851\pm0.013$    
\\ \\

\hline
\end{tabular}
\vspace{0.5cm}
\begin{tabular}{l|ccc}
Solubility & MPR & ECFP-KRR & GYLM-KRR            \\
prediction & (Ref.~\cite{lee2017optimal}) & (BenchML) & (BenchML)    \\
\hline
MAE   & 0.61             & $0.54 \pm 0.02$          & $0.430 \pm 0.003 $              \\
$R^2$ & 0.85             & $0.87 \pm 0.01$          & $0.908 \pm 0.003 $                
\end{tabular}
\caption{
Performance baseline correction of literature results using BenchML models. Top: Classification of ligands into ``actives'' and ``inactives'' based on random 90\%:10\% train:test splits. Shown is the comparison of ROC-AUCs measured for three 2D-fingerprint-based architectures: Random Matrix Discriminant (RMD), Support Vector Machine (SVM) as reported by Lee et al., and a standard SVM from the BenchML library. The datasets are taken from Lee {\it et al.}~\cite{lee2019ligand}. Bottom: Prediction of log solubility on the ESOL dataset~\cite{delaney2004esol}. MPR denotes the Marchenko-Pastur regression model by Lee {\it et al.}~\cite{lee2017optimal}, which is based on topological fingerprints (ECFP6, among others). ECFP-KRR and GYLM-KRR denote two kernel-ridge regressors from the standard BenchML model library. The metrics correspond to the models' test performance at a training fraction of 90\%. }
\label{tab:goofys}
\end{table}

 \begin{figure*}[hbt]
 \centering
 \includegraphics[width=0.9\textwidth]{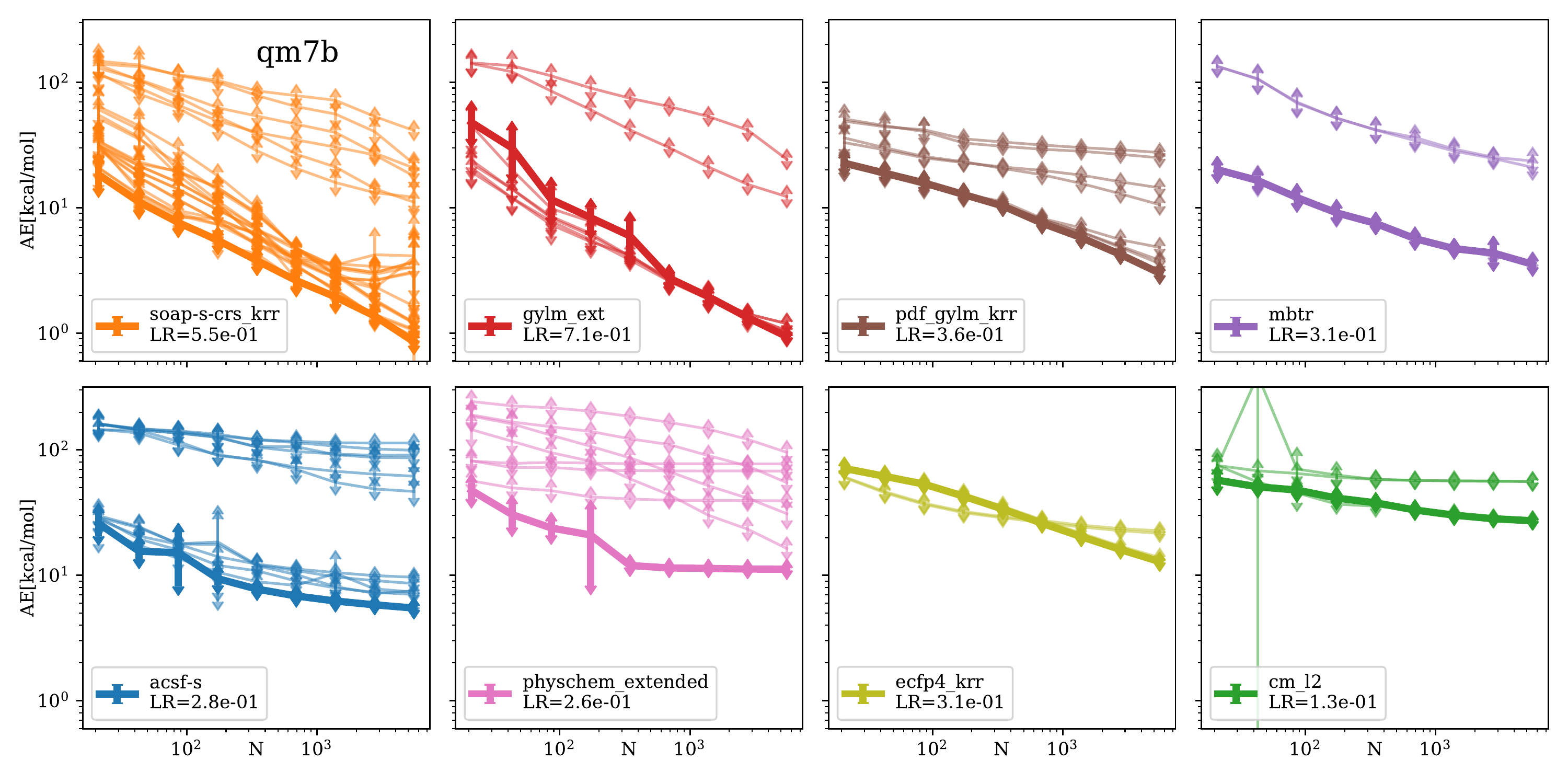}
\caption{
The learning curves (LCs) of the different models for the qm7b dataset. Each panel shows the test RMSE results for predicting atomization energy (AE) using the same representation, but different sets of hyperparameters,
Each thin line shows the LC of a specific model.
The LCs of the ``best-in-class'' models are highlighted using thick lines.
For clarity, we only show the legends for the best model for each representation.
The learning rate (LR) in the legends is defined as the slope of the learning curve of the best model in each panel.
}
\label{fig:qm7b-lc}
\end{figure*}

\subsection{BenchML in practice}

BenchML is focused on straightforward customization. New data transforms (listing~\ref{lst:descriptor}) are automatically registered upon import, and new ML pipelines (listing~\ref{lst:module}) can be added to the BenchML model library for immediate use. The default library contains many dozens of prebuilt models that can be applied quickly to new datasets. Some of these models are intended to serve as sensible baselines against which literature results can be compared.

The models benchmarked in this study are all part of this BenchML model library. The pipelines defined therein can be invoked from the command line or imported into a custom python script should this be desired. The models are tagged in a way that allows the user to run only a subset against a particular dataset. To give an example, the command
\begin{Verbatim}[fontsize=\small]
    bml --models "acsf.*" --mode benchmark \
        --meta qm7b_meta.json
\end{Verbatim}
would benchmark all models derived from the ACSF representation against the qm7b dataset, referenced here via its metadata file. BenchML uses these metadata files to specify raw file paths, provide instructions for the train-test splitting procedure, as well as convey certain prior information, which may be used by the model to intelligently select some of its hyperparameters. Listing~\ref{lst:metadata} exemplifies the metadata format in full detail. Whereas some of the metadata fields (such as the list of atomic elements) serves a mere practical purpose in that it informs the model about aspects of the dataset that cannot always be adequately inferred from a single training subset, other fields, in particular the ``scaling'' attribute provided for each target, assist the model in taking shortcuts through the hyperparameter search.

If, for example, an additive (extensive) property such as an energy is to be regressed using an intensive representation (such as a topological molecular fingerprint), then a sizable performance boost can be gained by first normalizing the target by molecular size, regressing the resulting intensive property, and then multiplying again by size. Clearly this kind of standardization could be made part of the preprocessing of the data, except that the appropriate preprocessing procedure will typically depend on the model architecture, as well as that externally performed preprocessing interferes with an end-to-end philosophy that is often desirable from a deployment point-of-view. Other metadata fields are designed to check scope of applicability, by indicating whether the data is amenable to SMILES representations, or whether the objective is classification or regression. Finally, the metadata also specify training-test splits: Appropriate testing procedures will often vary from one dataset to another. Common approaches are random, chronological and group-based splitting. The example in listing~\ref{lst:metadata} uses a sequential splitting mode for learning curve generation, where each training fraction $f \in [0.1, 0.9]$ is repeated $n = \lfloor \sqrt{4/f(1-f)} \rfloor$ times in order to ensure adequate sampling.

We now give a very brief demonstration how the BenchML model library can help us to quickly and easily gauge the predictive performance achieved by a novel ML technique. Clearly baseline selection is a key issue in evaluating the merit of a newly published technique. Even when comparing with sensible baseline methods, a comparison can be flawed if those baselines have not been carefully trained, or if hyperparameter tuning has been skipped. The example we use here stems from the domain of ligand-protein activity predictions. An earlier study~\cite{lee2019ligand} has found that a {\it Random Matrix Discriminant} (RMD) displayed superior performance in classifying compounds into actives and inactives for a set of five protein targets. Among one of the baseline models that was drastically outperformed by the RMD was a topological SVM based on ECFP fingerprints. Having downloading the underlying dataset, we can in three simple steps benchmark the BenchML version of that SVM against the literature data:
\begin{Verbatim}[fontsize=\small]
    binput --from_csv activity.csv \ 
           --output activity.xyz
    bmeta --extxyz activity.xyz \ 
          --meta input.json
    bml --mode benchmark --meta input.json \ 
        --models "ecfp_svm_class" 
\end{Verbatim}
Here the first command converts the csv into an extended-xyz input file; the second command generates a metadata file; the third command invokes the benchmark. The results, summarized in table~\ref{tab:goofys}, clearly show that the conceptually simpler SVM in fact outperforms the RMD by a small but significant margin, contrary to the authors' original claim.

As another example where a novel method is easily outperformed by simpler approaches, we point to the prediction of log solubility on the ESOL dataset using a regression model with a Marchenko-Pastur filtering step (a variant of principal-component analysis~\cite{lee2017optimal}). Again, a simple topological kernel outperforms the authors' original model, achieving a reduction in mean absolute error (MAE) from 0.61 to 0.54. A geometric kernel pushes this even further down to an MAE of 0.43 (see Table~\ref{tab:goofys}, bottom).

These examples highlight that the development of novel performant methods is hard and becoming even harder as the field reaches maturity. Detailed benchmarking is thus an increasingly important tool that helps us to build confidence in the merit of new ideas and approaches.

  \begin{figure*}[hbt]
  \centering
\includegraphics[width=0.95\textwidth]{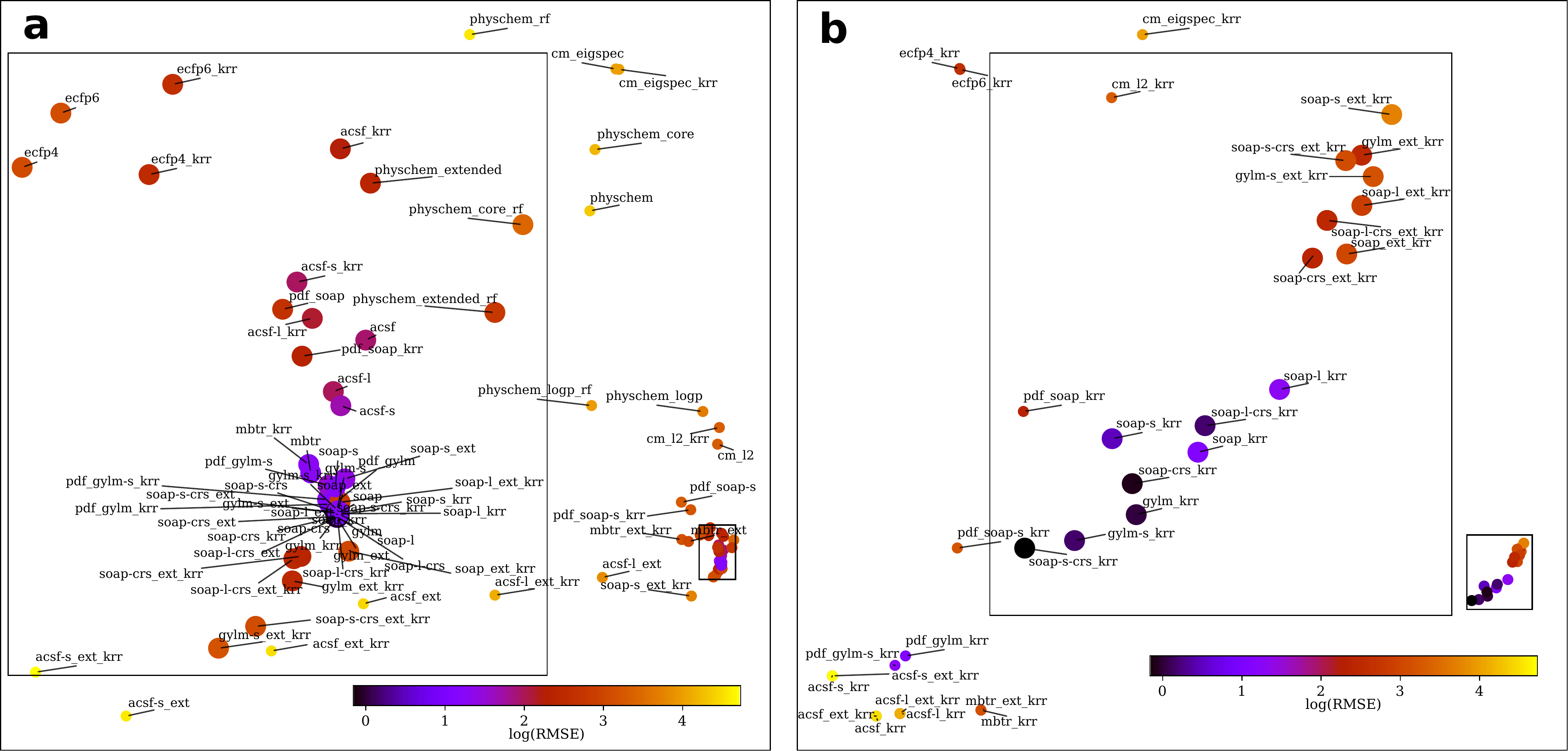}
\caption{
Similarity among models (a) and representations (b).
Panel a: KPCA projection of the kernel matrix $\bK^\mathrm{model}$ that measures the similarity between pairs of sets of model predictions (e.g. $k^\mathrm{model}(\mathcal{A},\mathcal{B})=R(\by_\mathrm{test}^{\mathcal{A}},\by_\mathrm{test}^{\mathcal{B}})$) on a (possibly unlabeled) test set $\bX$.
Each point denoting one model is colored according to the RMSE of the model on the same test set.
Panel b:
KPCA projection of the kernel matrix $\bK^\mathrm{desc}$ that measures the similarity between pairs of representations (e.g. $k^\mathrm{desc}(\Phi,\Phi')=R(\bK^{\Phi}(\bX^{(M)}), \bK^{\Phi'}(\bX^{(M)}))$), computed from $100$ randomly selected samples from the qm7b dataset. Each point is annotated by the label of the model it corresponds to; the points are coloured according to RMSE of the KRR fits. 
In each panel, the inset is a zoom-in of the area enclosed by the small rectangle.
}
\label{fig:qm7b}
\end{figure*}

\section{A specific application with detailed analysis~\label{sec:analysis}}

We here illustrate several analysis endpoints that allow us to study model performance in absolute and relative terms, as based on the output of a BenchML benchmark. This example is based on the qm7b dataset (a molecular dataset of molecular properties, in particular DFT-based atomization energies) and considers most of the available representations implemented to date. A more comprehensive benchmark spanning various datasets and models will be presented in section~\ref{sec:examples}.

We first focus on learning curves (LCs) for the regression of atomization energies of qm7b structures. LCs simulate model performance across multiple training regimes (from low to dense data) and are therefore a rigorous way of assessing model quality. These learning curves can be easily constructed and visualised starting from the output of a benchmark via
\begin{Verbatim}[fontsize=\small]
    bplot --input output.json.gz --output lcs.pdf
\end{Verbatim}

Fig.~\ref{fig:qm7b-lc} shows LCs generated accordingly for eight model families, with each panel grouping models according to individual branches of the ``phylogenetic'' tree from Fig.~\ref{fig:representations}. The panels are arranged such that the peak performance (i.e., the performance achieved by the best model within each family) decreases from top-left to bottom-right. Notice that for each representation, there is significant spread in performance across the family members (each of which corresponds to a different pooling rule, regression technique, as well as descriptor-specific hyperparameter settings). The spread affects both mean performance and learning efficiency and thus highlights the challenge in performing an objective benchmark, as even minor misalignment and poor choices in how a representation is embedded in an ML model can potentially ruin that model's performance.

\paragraph{Model-model error correlation}

Benchmarks typically focus primarily on estimating relative model performance -- as is achieved, for example, by comparing learning curves. A more fine-grained analysis is however needed to understand how models relate to each other on a mechanistic level -- i.e., how their predictions and their errors correlate on a sample-by-sample basis. This type of analysis can inform future method development, by shedding light on model bias and failure modes. Furthermore, we should be able to use this relatedness between models to construct ensembles of models that yield robust low-variance estimators by compensating for outlier predictions made by their individual members.

We first study model similarity based on their ability to rank samples of a test set according to their target values. Take two regression models $\mathcal{A}$ and $\mathcal{B}$. We denote their predictions on a test set (which is used neither during the hyperparameter search nor training) $\by_\mathrm{test}^{\mathcal{A}} = f^{\mathcal{A}}(\bX_\mathrm{test})$ and $\by_\mathrm{test}^{\mathcal{B}}= f^{\mathcal{B}}(\bX_\mathrm{test})$, respectively. We quantify the similarity $k^\mathrm{model}(\mathcal{A},\mathcal{B})$ among the set of predictions via their Spearman's rank correlation coefficient $R \in [-1, 1]$ between $\by_\mathrm{test}^{\mathcal{A}}$ and $\by_\mathrm{test}^{\mathcal{B}}$. Invariant with respect to target scale, domain, and to some degree, distribution, the rank correlation is attractive in that it allows us to aggregate similarity statistics across several datasets in a balanced way. Note that, even though $R$ can be negative, in practice, even disparate models achieve $R(\by_\mathrm{test}^{\mathcal{A}},\by_\mathrm{test}^{\mathcal{B}}) \gg 0$. 

We compute $R(\by_\mathrm{test}^{\mathcal{A}},\by_\mathrm{test}^{\mathcal{B}})$ for every ML model pair, and thus obtain an $n^\mathrm{model} \times n^\mathrm{model}$ kernel matrix $\bK^\mathrm{model}$, where, for the example of the qm7b dataset, $n^\mathrm{model} = 72$. Notice that the actual labels of the test set $\by_\mathrm{test}$ were not needed in the construction of $\bK^\mathrm{model}$. In fact, evaluating the correlation matrix on the true errors of the predictions (which additionally require the data labels) results in a very similar kernel matrix.

To visualize $\bK^\mathrm{model}$, we use kernel principal component analysis (KPCA) to construct the two-dimensional map shown in Fig.~\ref{fig:qm7b}a. Each point on the map is annotated with the label of the model it corresponds to, and is coloured by its RMSE measured on the test set. Models that use the same representation tend to be grouped close together on the KPCA map, next to having similar test errors. For example, the two topological representations, ECFP4 and ECFP6, that differ only with respect to their topological diameter (4 vs 6), form their own cluster that is locally well separated from the other models. Meanwhile, models using the ACSF, SOAP, GYLM or MBTR descriptor -- all of which are atom-density-based representations -- form a dense cluster with additional substructure resulting from the pronounced similarity between SOAP and GYLM, and from ACSF being the ``outsider'' within this clique of models.

As a dominant characteristic of the map in Fig.~\ref{fig:qm7b}a, the ``good'' models with low RMSE cluster very closely together, whilst models that are farther from the center of mass of the map have progressively worse performance. In other words, the good models are all alike, while the bad models are all bad in their own way. This implies that, if the actual labels of the qm7b test set $\by_\mathrm{test}$ were not available, just by comparing the similarities between the model predictions on the unlabeled test set
one can make an educated guess as to which models are likely to have better accuracy for these test samples.

  \begin{figure*}[hbt]
  \centering
\includegraphics[width=0.99\textwidth]{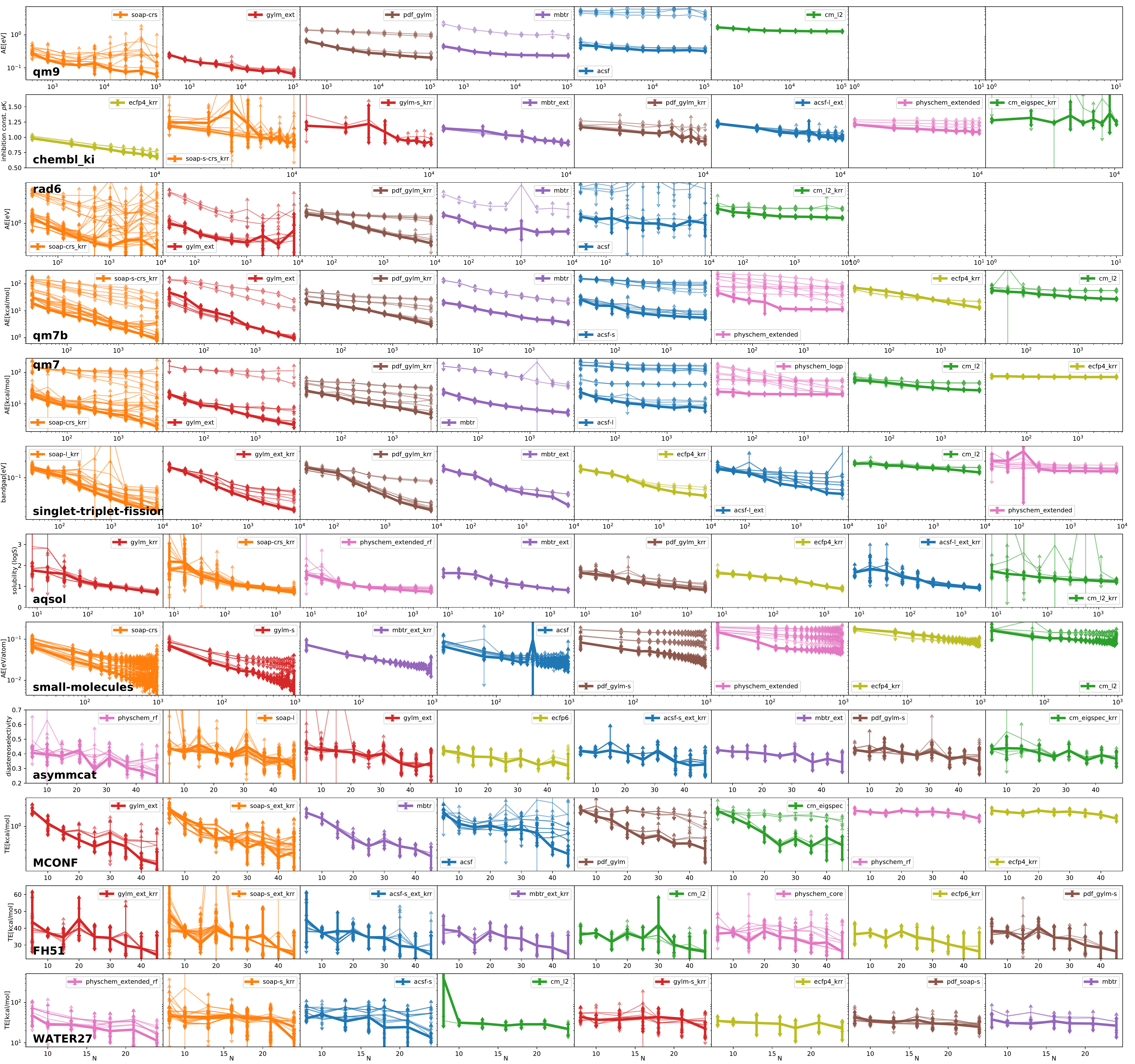}
\caption{
Learning curves (LCs) measured for the molecular datasets, showing the dependence of the test-set RMSE on the training set size $N$. 
AE: atomization energy; TE: total energy.
Each panel incorporates the results for one of the datasets described in Table~\ref{tab:DATABASES}. Each thin line corresponds to the LC of a specific model. For each dataset, we computed the LCs for 72 models. For clarity, we only show the curve label of the best model in each model family, with the corresponding LC of this ``best-in-class'' model highlighted with a thicker line width. The datasets are sorted from largest to smallest from top to bottom, and the model families are ranked from best to worst from left to right.
}
\label{fig:lc-molecule}
\end{figure*}

  \begin{figure*}[hbt]
\includegraphics[width=0.99\textwidth]{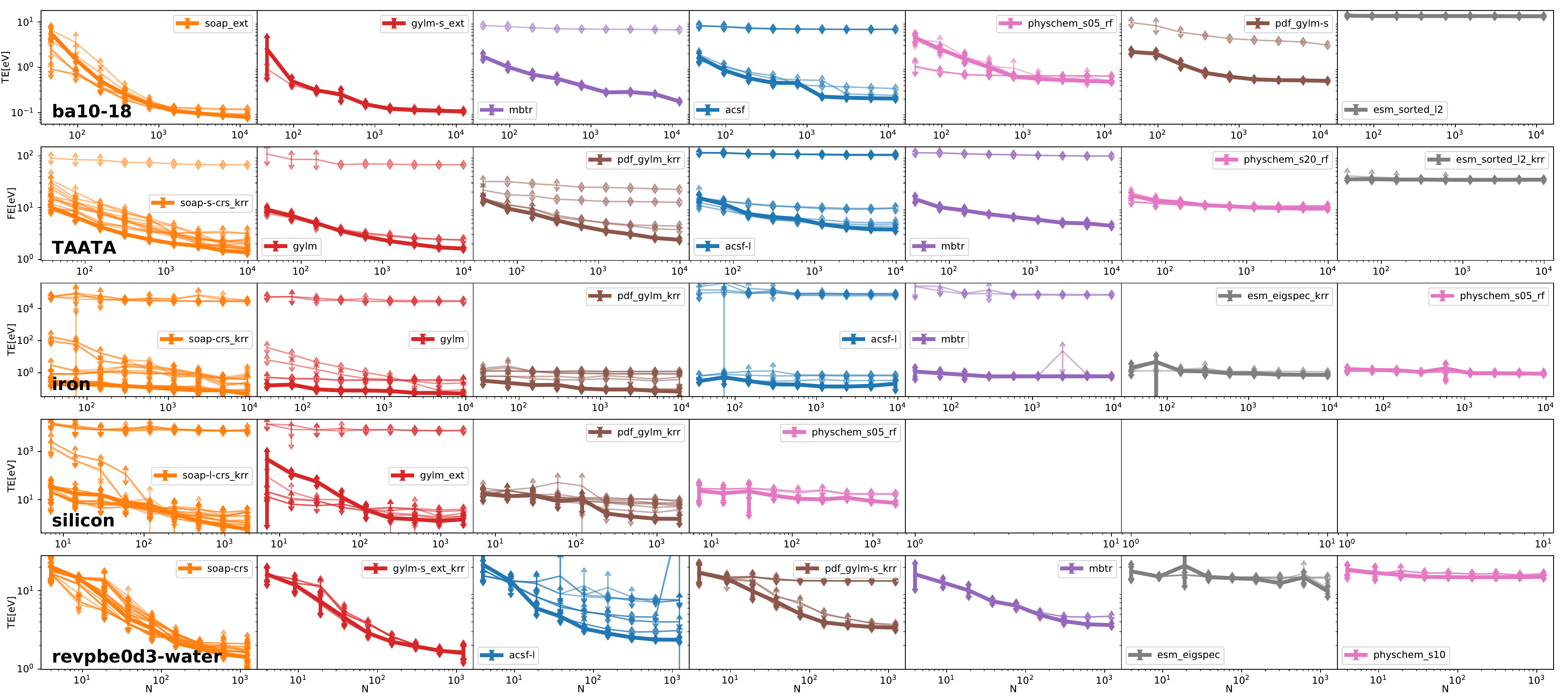}

\caption{
Learning curves (LCs) measured for the bulk-material datasets, showing the dependence of the test-set RMSE on the training set size $N$. 
FE: formation energy; TE: total energy.
Each panel incorporates the results for one of the bulk datasets described in Table~\ref{tab:DATABASES}. Each thin line corresponds to the LC of a specific model. For each dataset, we computed the LCs for 70 models. For clarity, we only show the curve label of the best model in each model family, with the corresponding LC of this ``best-in-class'' model highlighted with a thicker line width. The datasets are sorted from largest to smallest from top to bottom, and the model families are ranked from best to worst from left to right.
}
\label{fig:lc-bulk}
\end{figure*}\paragraph{Model-model feature-space correlation}

We next investigate model similarity via their respective feature spaces. Clearly models based on similar representations should yield similar predictions, whilst the converse -- similar predictions implying similar representations -- is less certain. In this benchmark, the regressors are simple linear regression or kernel ridge regression models, such that correlations in the feature spaces are clearly expected to carry over to the output layer of a model. Directly comparing different representations is not entirely straightforward, as they will typically differ in terms of both dimension and domain. We therefore cast each representation $\Phi$ into a kernel matrix $\bK^{\Phi}$ of a fixed size $M\times M$, and base the comparison on this dual-space representation. For a given dataset, we randomly select $M$ samples $\bX^{(M)}=(\bx_1,\ldots,\bx_M)^T$ and compute the kernel matrix between these $M$ samples using the kernel
\begin{equation}
    \bK^{\Phi}(\bX^{(M)}) = \Phi(\bX^{(M)}) \left[\Phi(\bX^{(M)})\right]^T.
\end{equation}
This dot-product kernel matrix is the same as used in the kernel-ridge regressors of the benchmarked models (Fig.~\ref{fig:model_phylogeny}. We measure the similarity $k^\mathrm{desc}(\Phi,\Phi')$ between pairs of representations $\Phi$ and $\Phi'$ by calculating their Spearman's rank correlation coefficient $R$ from the flattened kernel matrices $\bK^{\Phi}$ and $\bK^{\Phi'}$. 

To visualize $\bK^\mathrm{desc}$, we use KPCA to produce a two-dimensional map as shown in Fig.~\ref{fig:qm7b}b for qm7b dataset. The axes of the KPCA map are seen to be correlated with the measured test-set RMSE of the models. Once again, models with similar performance remain close on the map.
These observations are reminiscent of Fig.~\ref{fig:qm7b}a: We stress, however, that the construction of the KPCA map in Fig.~\ref{fig:qm7b}b does not rely on fitting of the models. This means that one can anticipate performance clusters among the models from their induced kernel space -- indicating that within the context of this benchmark representations and predictions are very much linked together, and that, importantly, drawing conclusions regarding the merits and drawbacks of different representations is justified.

\begin{table*}[htp]
    \caption{ML datasets in computational chemistry that formed part of the benchmark, ordered from largest to smallest.}
    \centering
    \renewcommand{\arraystretch}{1.5}
    {\scriptsize
    \begin{tabular}{c c p{12cm}}
    \hline
        Database & size & Description\\
        \hline
        \textbf{Molecular} & & \\
        qm9 & 133,885 & Hybrid DFT derived structures and properties of drug-like molecules with up to nine heavy atoms (C, O, N, or F). Initial configurations are taken from a subset of the GDB-17 data-set~\cite{GDB17}. Properties have been calculated for all molecules, including: energies of atomization, as well as other electronic and thermodynamic properties. Here we fit the atomization energies. \\
        
        chembl\_ki & 11,444 & Binding affinity data (inhibition constants $K_i$) for seven selected protein targets (5HTT, ADA3, BACE, GR, HERG, HIV1PR, VEGFR, thrombin). \\
                
        rad6 & 10,712 & Molecules consist of H,C,O elements up to 6 heavy atoms. The database comprises 9179 radical fragments and 1533 closed shell molecules~\cite{stocker2020machine}. Here we fit the atomization energies.\\
        
        singlet-triplet-fission & 9,919 & Singlet and triplet band gaps of indolonaphthyridine thiophene materials calculated using time-dependent density functional theory (TD-DFT) on DFT-optimized structures~\cite{Fallon2019}. %
        Here we fit the singlet band gap. \\
        
        qm7b & 7,211 & Small molecules with up to 7 heavy atoms, an extension for the qm7 dataset with additional properties~\cite{montavon2013machine}. 
        Here we fit the atomization energies.\\
        
        qm7 & 7,165 & Small molecules selected from GDB-13 (a database of nearly 1 billion stable and synthetically accessible organic molecules) containing up to 7 heavy atoms C, N, O, and S~\cite{rupp2012fast}. Here we fit the atomization energies.\\

        aqsol & 2,906 & Aqueous solubility dataset, incorporating the ESOL dataset by Delaney~\cite{delaney2004esol} and public domain data. \\

        small-molecules & 985 & Selected from qm7b set using the farthest point sampling algorithm. \\
            
        asymmcat & 53 & Asymmetric hydrogenation: dependence of the diastereoselectivity on ligand structure (Poelking, Amar et al.~\cite{poelking2019noisy}) \\

        FH51 & 51 & Reference reaction energies of 51 reactions for small molecules~\cite{zhao2005benchmark, friedrich2013incremental}. From GMTKN55~\cite{Goerigk2017}(a database for general main group thermochemistry, kinetics, and non-covalent interactions). \\
        
        MCONF & 51 & Reaction energies of melatonin conformers~\cite{Fogueri2013}. From GMTKN55~\cite{Goerigk2017}. \\
        
        WATER27 & 27  & Energies of 27 water clusters, up to 20 water monomers~\cite{bryantsev2009evaluation,anacker2014new}. From GMTKN55~\cite{Goerigk2017}. \\
        
        \hline
        \textbf{Bulk} & & \\
        
        ba10-18 & 15,950 & Energies for 10 binary alloys (AgCu, AlFe, AlMg, AlNi, AlTi, CoNi, CuFe, CuNi, FeV, NbNi) with 10 different species and all possible face-centered cubic (fcc), body-centered cubic (bcc) and hexagonal close-paced (hcp) structures up to 8 atoms in the unit cell~\cite{nyshadham2019machine}. \\
        
        TAATA & 12,823 & Consists of DFT predicted crystal structures and formation energies from the Ti-Zn-N, Zr-Zn-N, and Hf-Zn-N phase diagrams.~\cite{tholander2016strong} \\
        
        iron & 12,193  & The training set of a GAP ML potential for bcc ferromagnetic iron, which contains configurations with different number of atoms ranging from 1-130~\cite{Dragoni2018}.\\

        silicon & 2,475 & Training set of the  a Gaussian approximation potential for silicon~\cite{bartok2018machine}. \\
        
        revpbe0d3-water & 1,593 & The training set of a ML potential for bulk liquid water~\cite{cheng2019ab}. Contains 1,593 configurations of 64 molecules each. \\

        \hline
    \end{tabular}}
    \label{tab:DATABASES}
\end{table*}

\section{Examples applications of BenchML on popular chemical datasets~\label{sec:examples}}

We applied BenchML to a number of chemical datasets as summarized in Table~\ref{tab:DATABASES}. These datasets are classified into two categories, {\it molecular} and {\it bulk}, with bulk datasets consisting of periodic (amorphous, disordered and crystalline) structures as opposed to isolated molecules or clusters. All the datasets as well as their metadata specification are included in the Supplementary Data. 

For the molecular datasets, the representations (see Fig.~\ref{fig:model_phylogeny}) considered in the benchmark include the physicochemical (PhysChem), ECFP, MBTR, GYLM, ACSF, SOAP, and Coulomb matrix representations. As the PDF contraction of SOAP or GYLM change the nature of the representations in a fundemental way, we treat them as a separate class. Note that for datasets that contain dissociated molecules with broken or dangling bonds (qm9 and rad6), representations that rely on a healthy chemical topology (i.e., PhysChem and ECFP) are excluded. 

The learning curves for the molecular datasets are provided in Fig.~\ref{fig:lc-molecule}. For each representation, LCs corresponding to its model variations (which combine descriptor subtypes with different pooling and regression rules) are displayed separately. Additionally, the LC corresponding to the ``best-in-class'' model is highlighted with a thicker line width.

For the bulk datasets we considered six families of models adapted from the PhysChem, MBTR, GYLM, ACSF, SOAP, and the Ewald matrix representation. The corresponding LCs are shown in Fig.~\ref{fig:lc-bulk}. Note that, as indicated in Fig.~\ref{fig:model_phylogeny}, the PhysChem representation for these periodic systems is no longer based on physicochemical molecular features, but a smaller set of SISSO atomic features~\cite{luca2018sisso}.

As previously observed in the qm7b case study, there is considerable variation in the learning outcome even within each model family. This variation is most pronounced for the 3D geometric representations, where length-scale hyperparameters appear to be particularly decisive -- once again highlighting the need for appropriate hyperparameter selection. It is worth noting that, for ACSF, besides the length-scale hyperparameters discussed above, there is the option to fine-tune the exact specifications of the symmetry functions. The models studied here do not exploit this option. Instead, the function parameters are selected from a uniform grid, which may explain why the ACSF representation (which was originally designed for use in neural-network architectures) underperforms in this benchmark. Even though this poor performance could thus potentially be remedied using more sophisticated heuristics for selecting the basis functions, neither standard grid nor Bayesian hyperparameter optimization on a single-task level are particularly well suited for this purpose, due to the large size of the parameter space. It is therefore naturally appealing when a representation with lower parametric complexity manages to perform well even without complicated heuristics and routines for selecting these parameters (as accomplished, e.g., by SOAP).

Note that in both Fig.~\ref{fig:lc-molecule} and Fig.~\ref{fig:lc-bulk}, the datasets are sorted from {\it largest} to {\it smallest}, as quantified by the number of distinct structures contained therein; additionally, the model families are ranked from {\it best} (left-most column) to {\it worst} (right-most column), as judged from the RMSE achieved at the largest training fraction by the best-in-class model of each representation. Systematic trends around the learning outcomes with respect to dataset type and size are thus easy to discern. In the high-data-volume regime, SOAP is consistently the top-performer, whereas with the smallest datasets, the 1D physicochemical representations prevail. This robust performance of 1D representations for smaller datasets is well-known, and rationalized by their low dimension and high information density. 

A notable exception to this trend is, however, the regression of binding affinities (chembl\_ki  dataset). Recall that we considered only simple pooling rules to derive a molecular representation from individual atomic vectors. As a result, the identity and characteristics of molecular subgroups are not necessarily well preserved, so that straightforward identification of the motifs that form key interactions with the protein is unlikely.  Detection of such motifs is more easily -- and virtually by design -- achieved by hashed topological fingerprints, which are geared towards substructure recognition, as reflected by the superior performance that this family of models achieves on this particular dataset. We point out, however, that pairing local-environment-based representations with more sophisticated (nonlinear) pooling rules significantly improves the performance of 3D descriptors, albeit at a significantly increased computational cost~\cite{bartok2017unifies,poelking2020meaningful}. 

Finally, among the 3D representations, we note that SOAP and GYLM perform quite similarly across several of the datasets included in this benchmark. Both of these representations are based on spherical harmonics, with GYLM furthermore adopting SOAP-type contractions to enforce rotational invariance. Even though there are key differences in how these representations achieve regularization (i.e., dampening of high-frequency structural features) and prioritise close over far neighbours, their similarity in performance points to the merits of the power spectrum in crafting expressive representations from basis-function expansions of the nuclear density. Interestingly, extending this power-spectrum to include non-local pairwise contractions in the form of Eq.~\ref{eq:pair_spectrum} does not appear to be remotely beneficial, as highlighted by the poor performance of the PDF family of models. Local pooling rules thus remain particularly attractive for regressing additive properties (which are often of a surprisingly local nature), in that they naturally avoid the distraction presented by irrelevant long-range structural correlations and global conformational flexibility.

\section{Conclusions}

The tremendous progress that has been made over recent years in the area of chemical ML has provided us with a wealth of chemical representations and predictive models. As a result, benchmarking is becoming ever more important in order to evaluate the benefits of new approaches and, in doing so, differentiate anecdotal from statistically relevant progress. Here we presented BenchML -- a general, extensible pipelining framework designed with both model validation and deployment in mind. Intended to provide a simple route how to make large-scale benchmarking against multiple datasets part of the method development process, the framework also allows for integrating performant models with confidence prediction and attribution -- both of which are common prerequisites for successful model deployment.

Our benchmark highlighted that there is significant complexity in how representations can or should be embedded even in very simple ridge and kernel ridge regressors, with significant variation in performance observed within individual model families. Casting ML models as pipelines thus comes with the key benefit that even complex approaches that embed a given representation into a predictive architecture can be explored concurrently and with ease. The layout of the pipeline, the parameters of the pre- and post-processing stages (such as pooling and reduction rules), as well as the parameters of the representations themselves can be tuned either automatically or preset by the modeller.

Learning curves that we recorded for a variety of datasets illustrated the relative merits of atomic, molecular and bulk representations. KPCA approaches furthermore allowed us to visualise relationships between models based on their error and feature-space correlations. Geometric representations, in particular SOAP, excelled at regressing additive properties for high-volume datasets. Topological fingerprints performed well in predicting non-additive properties, as shown here for binding affinity modelling. Physicochemical representations dominated in ultra-low-data settings. Apart from the representation itself, the choice of pooling and contraction rules proved most important in determining the modelling outcome. The top-performing representations, SOAP and GYLM, are fundamentally related in that they both use the power spectrum to construct their atomic descriptions. Deviations from this local, pairwise contraction rule proved harmful, as indicated by the performance downturn of models that limit the number of element-element cross-channels or that adopted non-local contractions.

In this vein, we hope that large-scale benchmarks can be used not only to verify the merit of novel methods and representations, but also to further our mechanistic understanding of atomic and molecular representations in a way that over time allows for targeted improvement of their form and function.\\

\textbf{Acknowledgements}
CP acknowledges funding from Astex through the Sustaining Innovation Program under the Milner Consortium. BC acknowledges resources provided by the Cambridge Tier-2 system operated by the University of Cambridge Research Computing Service funded by EPSRC Tier-2 capital grant EP/P020259/1. FAF acknowledges funding from the Swiss National Science Foundation (Grant No. P2BSP2\_191736). \\

\textbf{Data availability}
The datasets used for this study are available at \url{https://github.com/BingqingCheng/linear-regression-benchmarks}.\\

\textbf{Code availability}
The BenchML framework, its documentation and source code can be accessed at \url{https://github.com/capoe/benchml}.\\

\begin{lstlisting}[language=Python,float=*,firstline=1,lastline=26,caption=Implementation of a RandomDescriptor transform (example only),captionpos=b,label={lst:descriptor}]
import benchml.transforms as btf
import numpy as np

class RandomDescriptor(btf.MapTransform):
    default_args = {
      "mean": 0.,
      "std_dev": +1.,
      "dim": None
    }
    req_args = {"dim",}                        # <- Required fields to be specified in args
    req_inputs = {"structures",}               # <- Required inputs to be specified in inputs
    allow_stream = {"X",}                      # <- Enables write access to the data stream
    stream_samples = {"X",}                    # <- Affects how the tensor is sliced during splits
    precompute = True                          # <- Flags the transform for precomputation

    def _map(self, inputs, stream):
        X = np.random.normal(    
            loc=self.args["mean"],
            scale=self.args["std_dev"],
            size=(
                len(inputs["structures"]), 
                self.args["dim"]
            )
        )
        stream.put("X", X)                     # <- Dumps X in the private stream of the transform
        
\end{lstlisting}
\begin{lstlisting}[language=Python,float=*,firstline=27,lastline=59,caption=Definition of a simple topological kernel regressor,captionpos=b,label={lst:module}]
import benchml.transforms as btf

model = btf.Module(
    transforms=[
        btf.StructureInput(
            tag="input"                        # <- Assigning a tag "input" allows us to refer
        ),                                     #    to the data content below as "input.<field>"
        btf.TopologicalFP(                            
            tag="descriptor",     
            inputs={"structures": "input.structures"}
        ),
        btf.KernelDot(
            tag="kernel",
            inputs={"X": "descriptor.X"}
        ),
        btf.KernelRidge(
            tag="regressor",
            args={"alpha": 1e-5, "power": 2},
            inputs={"K": "kernel.K", "y": "input.y"}
        )
    ],
    hyper=btf.BayesianHyper(
        btf.Hyper({
            "regressor.alpha": [-4, 4],
            "regressor.power": [1, 3]
        }),
        convert={
            "regressor.alpha": lambda a: 10**a  # <- Perform search in log-transformed space
        }  
    ),
    broadcast={ "meta": "input.meta" },         # <- Metadata is broadcast to all transforms
    outputs={ "y": "regressor.y" }              
)
\end{lstlisting}
\begin{lstlisting}[language=Python,float=*,firstline=62,lastline=94,caption=Metadata specification for the qm7b dataset,captionpos=b,label={lst:metadata}]
{
 "name": "QM7B",
 "datasets": [ "qm7b.xyz" ],
 "elements": [ "C", "Cl", "H", "N", "O", "S" ],      # <- Used by representations whose layout
 "has_smiles": True,                                 #    depends on the set of unique elements
 "periodic": False,
 "hypersplit": {
  "method": "random",
  "n_splits": 10,
  "train_fraction": 0.75
 },
 "splits": [
  {
   "method": "sequential",                           
   "train_fraction": np.linspace(0.1, 0.9, 9),       # <- Learning curve split sequence
   "repeat_fraction_fct":           
       lambda s,t,v,f: 2*int(1./(f*(1-f))**0.5)      # <- Determines how often a split is repeated
  }
 ],
 "targets": {
  "ae_pbe0": {                                       # <- Refers to field in qm7b.xyz metadata
   "task": "regression",
   "scaling": "additive",                            # <- "additive" vs "non-additive"
   "convert": "",                                    # <- Optional: conversion function
   "metrics": [ "mae", "rmse", "rhop", "r2" ]
  }
 }
}


\end{lstlisting}

\end{document}